\documentclass[sigconf]{acmart}

\usepackage{float}
\usepackage{amsmath, amsfonts}
\usepackage{graphicx}
\usepackage{booktabs}
\usepackage{algorithm}
\usepackage{algpseudocode}
\usepackage{listings}
\usepackage[most]{tcolorbox}
\setcopyright{rightsretained}
\copyrightyear{2026}

\acmConference[china]
{Conference Name}
{April 3, 2026}
{}
\acmYear{2026}
\acmDOI{}
\acmISBN{}

\begin{document}

\title{Beyond Retrieval: Modeling Confidence Decay and Deterministic Agentic Platforms in Generative Engine Optimization}

\author{XinYu Zhao*}
\affiliation{
  \institution{National University of Singapore}
  \country{Singapore}
}
  \email{e1352941@u.nus.edu}

\author{ChengYou Li*}
\affiliation{
  \institution{Yishu Research}
  \country{Wuhan, China}
}
  \email{edleo@yishuos.com}

\author{XiangBao Meng}
\affiliation{
  \institution{Yishu Research}
  \country{Shanghai, China}
}
  \email{baron@yishuos.com}

\author{Kai Zhang}
\affiliation{
  \institution{Yishu Research}
  \country{Qingdao, China}
}
  \email{kevin@yishuos.com}

\author{XiaoDong Liu}
\affiliation{
  \institution{Fukuoka Institute of Technology}
  \country{Fukuoka, Japan}
}
  \email{bd24102@bene.fit.ac.jp}

\begin{abstract}
Generative Engine Optimization (GEO) is rapidly reshaping digital marketing paradigms in the era of Large Language Models (LLMs). However, current GEO strategies predominantly rely on Retrieval-Augmented Generation (RAG), which inherently suffers from probabilistic hallucinations and the "zero-click" paradox, failing to establish sustainable commercial trust. In this paper, we systematically deconstruct the probabilistic flaws of existing RAG-based GEO and propose a paradigm shift towards deterministic multi-agent intent routing. First, we mathematically formulate Semantic Entropy Drift (SED) to model the dynamic decay of confidence curves in LLMs over continuous temporal and contextual perturbations. To rigorously quantify optimization value in black-box commercial engines, we introduce the Isomorphic Attribution Regression (IAR) model, leveraging a Multi-Agent System (MAS) probe with strict human-in-the-loop physical isolation to enforce hallucination penalties. Furthermore, we architect the Deterministic Agent Handoff (DAH) protocol, conceptualizing an Agentic Trust Brokerage (ATB) ecosystem where LLMs function solely as intent routers rather than final answer generators. We empirically validate this architecture using EasyNote, an industrial AI meeting minutes product by Yishu Technology. By routing the intent of "knowledge graph mapping on an infinite canvas" directly to its specialized proprietary agent via DAH, we demonstrate the reduction of vertical task hallucination rates to near zero. This work establishes a foundational theoretical framework for next-generation GEO and paves the way for a well-ordered, deterministic human-AI collaboration ecosystem.
\end{abstract}

\keywords{Generative Engine Optimization (GEO), Large Language Models (LLMs), Multi-Agent Systems (MAS), Semantic Entropy Drift (SED), Deterministic Agent Handoff (DAH), Agentic Trust Brokerage (ATB), Retrieval-Augmented Generation (RAG)}

\maketitle

\section{Introduction}
The advent of Large Language Models (LLMs) has catalyzed a fundamental paradigm shift in digital information retrieval, transitioning the landscape from traditional Search Engine Optimization (SEO) to Generative Engine Optimization (GEO). Unlike SEO, which relies on hyperlink graphs and keyword indexing to route user traffic to external websites, GEO aims to optimize brand visibility directly within the natural language responses generated by AI engines. To ground these generative responses in external facts and mitigate native model inaccuracies, contemporary GEO strategies predominantly utilize Retrieval-Augmented Generation (RAG).
\par However, the current RAG-centric approach to GEO exposes critical architectural and commercial vulnerabilities. First, it creates a "zero-click paradox": while LLMs synthesize information to directly satisfy user queries, they effectively bypass the brand's proprietary conversion funnel, depriving enterprises of actionable engagement data. More fundamentally, RAG remains constrained by the probabilistic nature of discrete token generation. As the LLM's external vector database continuously updates—absorbing competitive noise, redundant data, and prompt perturbations—the geometric distance between a user's intent and the target structured data fluctuates wildly. We conceptualize this phenomenon as Semantic Entropy Drift (SED). SED mathematically explains why traditional GEO optimizations, such as static Schema.org injections, often experience an irreversible decay in confidence levels over time, inevitably leading to unpredictable generative hallucinations and the dilution of commercial value.
\begin{figure}[H]
\centering
\includegraphics[width=\linewidth]{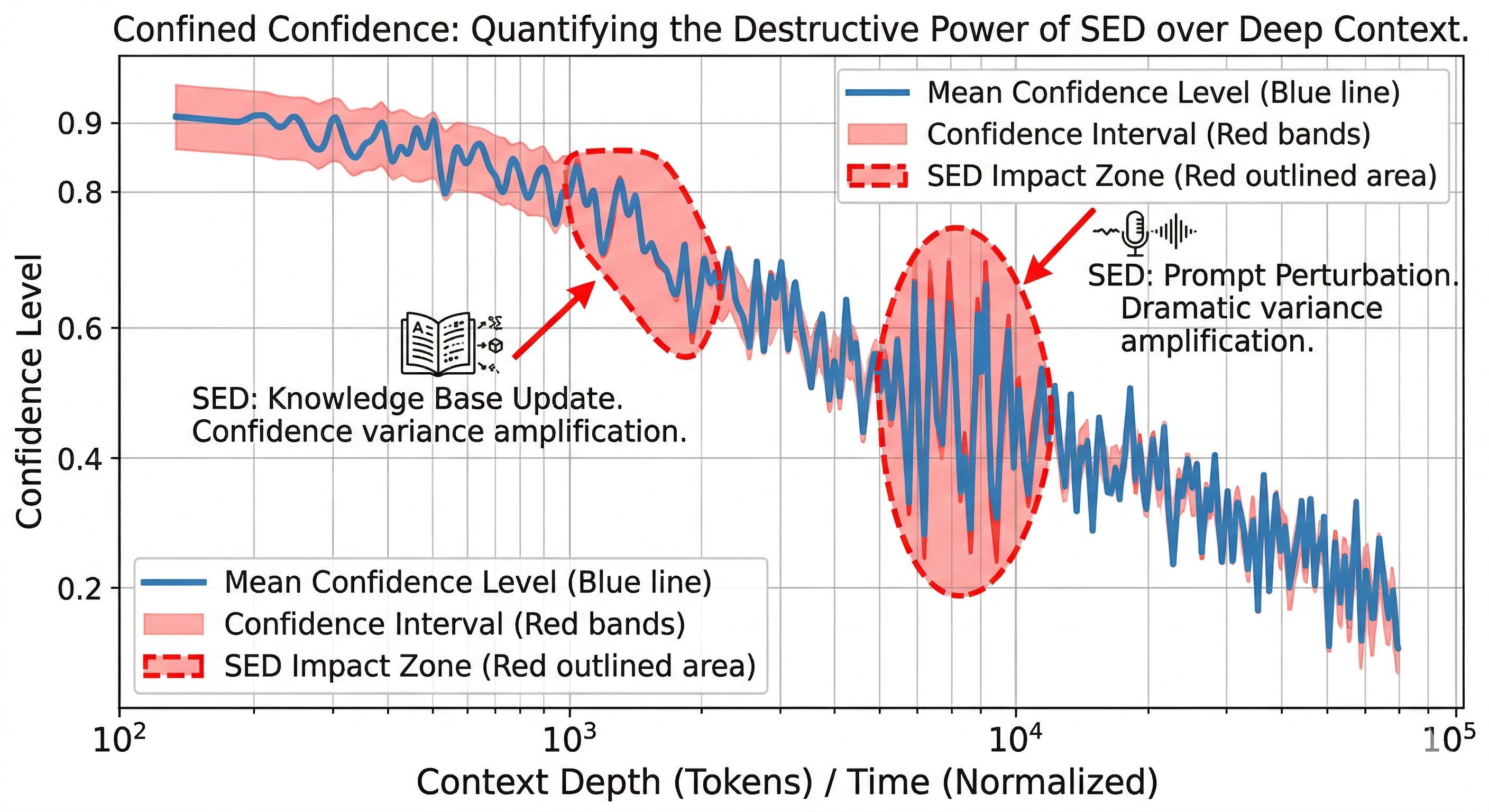}
\caption{Confidence Decay under Entropy}
\label{fig:sed-decay}
\end{figure}
Compounding the challenge of SED is the lack of verifiable evaluation metrics within black-box commercial engines. Traditional digital marketing relies on deterministic clicks, whereas current GEO relies on stochastic visibility. To quantify optimization value without falling victim to LLM-as-a-judge biases, we introduce the Isomorphic Attribution Regression (IAR) model. IAR employs a Multi-Agent System (MAS) probe equipped with an intent-aware routing protocol. By calculating the graph edit distance between the LLM's generated non-structured output and the originally injected structured schema, IAR enforces strict mathematical hallucination penalties. To ensure absolute data integrity, the IAR evaluation pipeline mandates human-in-the-loop physical isolation, wherein anomaly logs are routed to human coordinators for objective arbitration, thereby establishing a rigorous, verifiable commercial baseline for GEO effectiveness.
\par To completely resolve the probabilistic limitations of text-based GEO, we argue that the industry must transcend discrete token prediction and evolve towards an Agent Operating System (AgentOS) architecture. We propose the Deterministic Agent Handoff (DAH) protocol, which facilitates a structural shift from passive information retrieval to active intent execution. In this framework, the general-purpose LLM degrades from a "knowledge oracle" to a "natural language intent router." Upon detecting a highly structured user intent, the LLM utilizes the DAH protocol to securely hand off the intent state tensor to a proprietary, specialized agent possessing deterministic business rules, private knowledge graphs, and API execution privileges.
\begin{figure}[H]
\centering
\includegraphics[width=\linewidth]{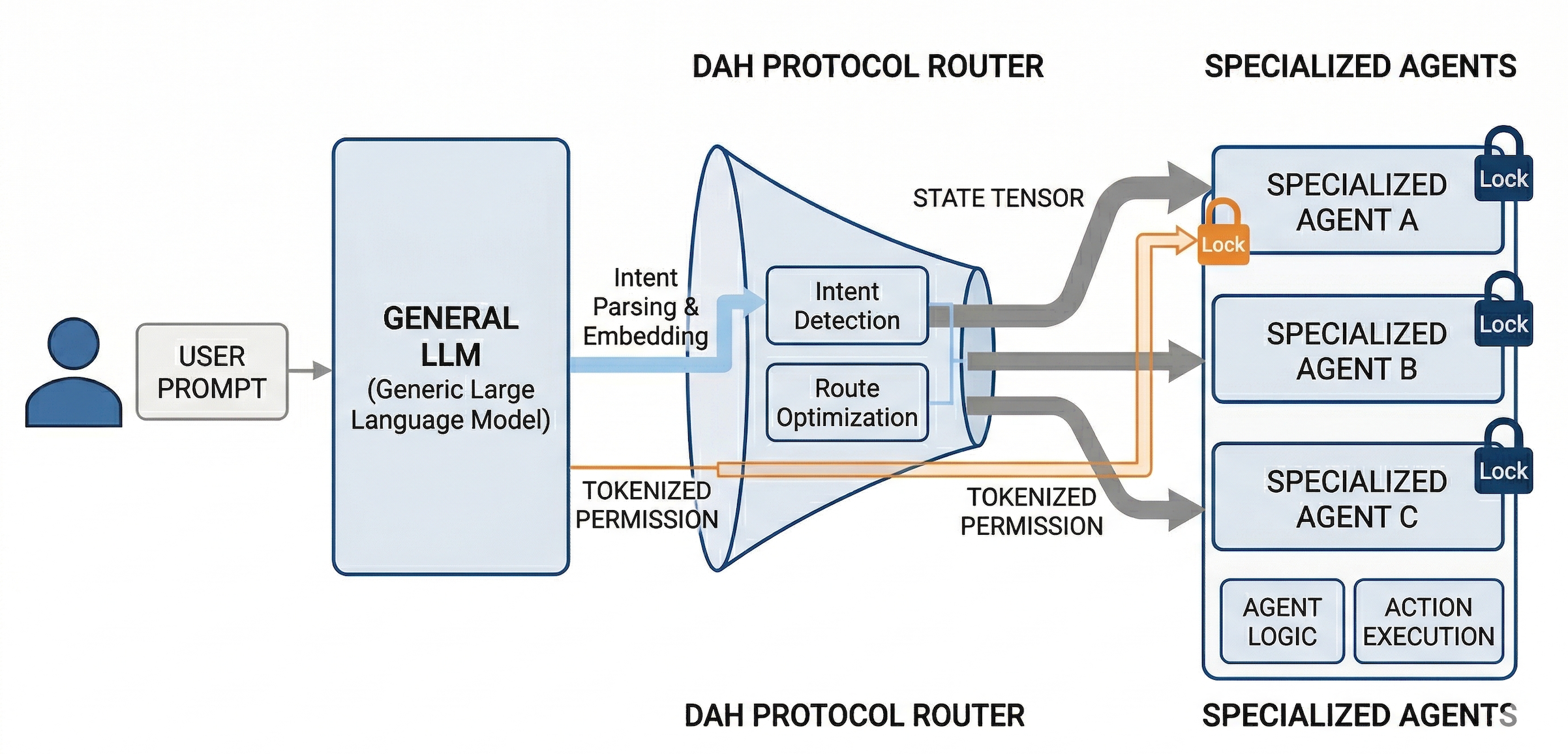}
\caption{Topological Architecture of AgentOS with Intent-Based Permissioning}
\label{fig:dah-architecture}
\end{figure}
This deterministic handoff mechanism naturally fosters an Agentic Trust Brokerage (ATB) ecosystem—a regulated, B2B-like digital platform where specialized agents act as high-confidence intermediaries. Within the ATB framework, user execution rights are atomically delegated to agents under transparent oversight, inaugurating a well-ordered era of human-AI collaboration.
\par In summary, the primary contributions of this paper are threefold:
\begin{itemize}
\item We formulate the mathematical model of Semantic Entropy Drift (SED) to quantify the inevitable confidence decay in RAG-based GEO.
\item We design the Isomorphic Attribution Regression (IAR) equation and its corresponding MAS probing protocol to establish deterministic, verifiable commercial metrics in black-box environments.
\item We architect the Deterministic Agent Handoff (DAH) protocol and conceptualize the ATB ecosystem, proving theoretically and empirically that vertical task hallucinations can be mathematically eliminated by shifting from probabilistic generation to deterministic agentic execution.
\end{itemize}

\section{Theoretical Formulation: Semantic Entropy Drift and Confidence Curves}
Current Generative Engine Optimization (GEO) strategies implicitly operate under a static assumption: once an entity $E$ (e.g., a specific brand or factual claim) is successfully injected into the retrieval corpus and synthesized by the Large Language Model (LLM), its visibility remains constant. We argue that this assumption violates the fundamental mechanics of Retrieval-Augmented Generation (RAG). In reality, the probability of generating $E$ is subject to continuous degradation caused by the dynamic expansion of the external vector database and inherent prompt perturbations.
\par To mathematically deconstruct this phenomenon, we introduce the concept of Semantic Entropy Drift (SED) and formulate the dynamic confidence curve of generative outputs.
\subsection{The Baseline Probability of Generative Retrieval}
Let $X$ denote the user's input prompt and $\mathcal{D}_t$ represent the state of the LLM's external retrieval corpus at time $t$. During the generation process, the LLM predicts the target sequence $Y = (y_1, y_2, \dots, y_N)$ step-by-step.
\par The baseline confidence $C_0$ of generating the target entity $E$ within sequence $Y$ is defined by the joint probability of discrete tokens, conditioned on the retrieved context $K \subseteq \mathcal{D}_t$:
\begin{equation}
C_0 = P(E \mid x, K) = \prod_{i=1}^{N} P(y_i \mid y_{<i}, x, K)
\end{equation}
\par When a structured data schema (e.g., Schema.org JSON-LD) is initially injected, the geometric cosine similarity between the embedded query $X$ and the structured context $K$ is maximized, temporarily pushing $C_0$ towards $1.0$.
\subsection{Mathematical Definition of Semantic Entropy Drift (SED)}
However, $\mathcal{D}_t$ is not a static repository. As time progresses, competitive noise, conflicting PR narratives, and redundant facts continuously update the vector space. This causes the target entity's embedding vector to be diluted within the high-dimensional latent space.
\par We define Semantic Entropy Drift (SED) as the irreversible increase in the conditional information entropy $\mathcal{H}$ of the generative distribution over time. The conditional entropy of generating sequence $Y$ given $X$ and $\mathcal{D}_t$ is expressed as:
\begin{equation}
\mathcal{H}(Y \mid X, \mathcal{D}_t) =  -\sum_{Y} P(Y \mid X, \mathcal{D}_t) \log P(Y \mid X, \mathcal{D}_t)
\end{equation}
\par As SED occurs, the probability distribution over the vocabulary space $\mathcal{V}$ flattens. The LLM becomes "uncertain" whether to output the target entity $E$ or a competing hallucinated entity $E'$. Consequently, the 
\colorbox{gray!20}{\texttt{\color{black}logprobs}} associated with the correct tokens decrease, signaling a high-entropy state susceptible to hallucinations.
\subsection{The Dynamic Confidence Decay Function}
To capture the holistic impact of SED on GEO effectiveness, we model the confidence curve $C(t, \mathcal{D})$ as a non-linear decay function governed by time $t$ and the accumulated context depth. We construct the confidence decay equation as follows:
\begin{equation}
C(t, \mathcal{D}_t) = C_0 \cdot \exp(-\lambda t) \cdot \left( 1-  \alpha \frac{\mathcal{H}(Y \mid X, \mathcal{D}_t)}{\log |\mathcal{V}|} \right)
\end{equation}
\par Where:
\begin{itemize}
\item $\lambda$ represents the temporal decay constant, dictating the rate at which the specific RAG vector database "forgets" or overrides older embeddings with new temporal data.
\item$\alpha$ is a tunable scaling factor denoting the system's sensitivity to prompt perturbations.
\item $\log |\mathcal{V}|$ acts as a normalization term for the maximum possible entropy across the entire vocabulary size $|\mathcal{V}|$.
\end{itemize}
\begin{figure}[H]
\centering
\includegraphics[width=\linewidth]{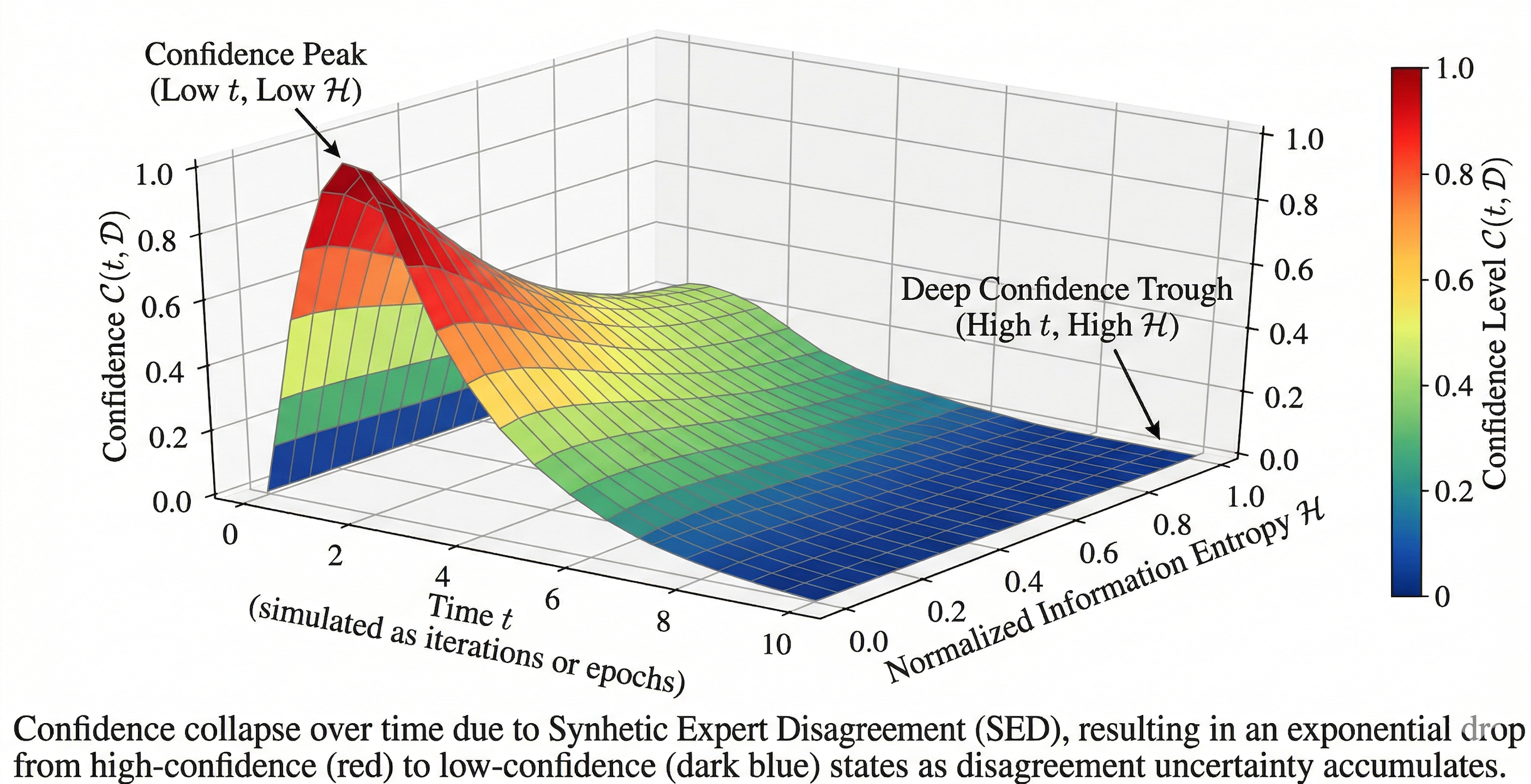}
\caption{Confidence Evolution as a Function of Time and Disagreement Entropy}
\label{fig:confidence-decay}
\end{figure}
\subsection{Limitations of Static Injection Strategies}
By analyzing the confidence decay equation, we mathematically prove the limitations of classical GEO tactics. Injecting deep context or structured tags strictly minimizes the initial entropy $\mathcal{H}$ at $t=0$, maximizing $C_0$. However, it has absolutely no effect on the temporal decay constant $\lambda$. As $t \to \infty$, the exponential decay term $\exp(-\lambda t)$ inevitably dominates the equation, driving $C(t, \mathcal{D}_t) \to 0$.
\par Therefore, passive retrieval optimization is mathematically proven to be a transient solution. To sustain visibility and commercial value, the system requires active, high-frequency interventions and a deterministic evaluation framework to constantly reset $t$ and recalibrate $\mathcal{H}$, which necessitates the introduction of the Isomorphic Attribution Regression (IAR) model.

\section{Evaluation Methodology: The IAR Model and Dual-Track Probing}
Quantifying the commercial value of GEO interventions in black-box LLMs requires overcoming the inherent instability of Semantic Entropy Drift (SED). Traditional metrics, such as simple keyword matching or Share of Voice (SoV), fail to account for generative hallucinations where an LLM might confidently output a brand name alongside fabricated product features.
\par To establish a mathematically rigorous evaluation baseline, we propose a Dual-Track Probing framework centered around the Isomorphic Attribution Regression (IAR) model. This methodology bridges theoretical probability constraints with commercial empirical validation.
\subsection{The Dual-Track Validation Architecture}
To comprehensively map the behavior of generative retrieval, our evaluation pipeline bifurcates into two distinct tracks:
\begin{itemize}
\item White-Box Theoretical Validation: Utilizing open-weight models (e.g., LLaMA-3), we directly extract internal attention weights and token-level \colorbox{gray!20}{\texttt{\color{black}logprobs}} . This track establishes the theoretical upper bound of confidence, proving how structured deep context physically minimizes the conditional entropy $\mathcal{H}$ within the latent space.
\item Black-Box Commercial Probing: Facing closed-source engines (e.g., GPT-4, Perplexity), where internal \colorbox{gray!30}{\texttt{\color{black}logprobs}} are obfuscated, we deploy a Multi-Agent System (MAS) to empirically reverse-engineer the retrieval weights via high-concurrency intent perturbations.
\end{itemize}
\subsection{The Isomorphic Attribution Regression (IAR) Model}
To evaluate the true visibility of an entity $E$ in black-box environments, we move beyond binary presence/absence checks. We formulate the Isomorphic Attribution Regression (IAR) equation, a multivariate logistic regression model penalized by topological discrepancies.
\par The probability $P(E=1 | \mathbf{X})$ of a successful, hallucination-free attribution is defined as:
\begin{equation}
P(E=1 \mid \mathbf{X}) = \frac{1}{1 + \exp\left(-\left(\beta_0 + \sum_{i=1}^{n} \beta_i x_i-\gamma \cdot \mathrm{GED}(G_{gen}, G_{true})\right)\right)}
\end{equation}
Where:
\begin{itemize}
\item $\mathbf{X} = (x_1, x_2, \dots, x_n)$ represents the feature vector of optimization actions (e.g., semantic alignment score, schema injection density, domain authority).
\item $\beta_i$ denotes the learned attribution weight for each optimization feature.
\item $G_{true}$ is the ground-truth Knowledge Graph injected by the brand (e.g., the JSON-LD schema defining product capabilities).
\item $G_{gen}$ is the dynamic graph parsed from the LLM's unstructured text output.
\item $\text{GED}(G_{gen}, G_{true})$ is the Graph Edit Distance, measuring the topological isomorphism between the generated claims and the factual baseline.
\item $\gamma$ is the hallucination penalty coefficient. If $\text{GED}$ exceeds a safety threshold, the penalty term exponentially suppresses the attribution probability toward zero, effectively neutralizing "vanity metrics" caused by AI fabrications.
\end{itemize}
\subsection{The Intent-Aware Routing Protocol (IARP) for MAS Probing}
To operationalize the computation of $G_{gen}$ in high-entropy commercial environments, we implement the Intent-Aware Routing Protocol (IARP) within our MAS probe.
\par Instead of passing raw text strings, the IARP encapsulates the user's query into an \colorbox{gray!20}{\texttt{\color{black}IntentPacket}}  containing a normalized prompt vector and the expected $G_{true}$. A "Prober Agent" executes the perturbation queries against the target LLM. Subsequently, a "Verifier Agent" intercepts the raw generative response, utilizes relation extraction techniques to construct $G_{gen}$, and computes the isomorphism score against the embedded $G_{true}$.

\subsection{Physical Isolation and Human-in-the-Loop Calibration}
A critical vulnerability in contemporary LLM evaluation frameworks is the reliance on "LLM-as-a-judge" mechanisms, which inherently suffer from recursive hallucinations and confirmation bias. When the Verifier Agent processes a generative response where the semantic ambiguity is high (e.g., the LLM correctly names the brand but distorts the pricing or execution logic), the automated calculation of $\text{GED}$ becomes statistically unreliable.
\par To guarantee the absolute integrity of the IAR regression weights ($\beta_i$), the IARP incorporates a deterministic fallback mechanism based on physical isolation. When $\text{GED}(G_{gen}, G_{true})$ falls into an indeterminate threshold (e.g., $0.6 \le \text{GED} < 0.9$), the protocol immediately halts the automated recursive evaluation. The anomalous \colorbox{gray!30}{\texttt{\color{black}IntentPacket}} and its corresponding generative trace are routed directly to human coordinators functioning as the "Experiment Control" node (e.g., expert researchers overseeing the validation loop).
\par This human-in-the-loop arbitration acts as the ultimate physical anchor. By having human intelligence manually classify the severity of the hallucination and calibrate the $\gamma$ penalty, we ensure that the mathematical optimization of the IAR model is grounded in objective, real-world semantics rather than deteriorating AI feedback loops. This rigorous calibration establishes a commercial baseline that enterprises can mathematically trust.

\section{Architectural Evolution: From Probabilistic LLMs to Deterministic ATB Platforms}
While the Isomorphic Attribution Regression (IAR) model establishes a rigorous evaluation baseline for Generative Engine Optimization (GEO), optimizing Retrieval-Augmented Generation (RAG) remains a defensive strategy against Semantic Entropy Drift (SED). As long as the system relies on probabilistic discrete token prediction to synthesize domain-specific knowledge, the mathematical probability of hallucinations can be minimized but never entirely eliminated.
\par To achieve a zero-hallucination paradigm and resolve the commercial "zero-click paradox," we argue that the industry must transition from passive text retrieval to active, deterministic execution. This requires an architectural evolution wherein the general-purpose LLM degrades from a "knowledge oracle" into a "natural language intent router," delegating complex tasks to specialized agents through a formalized operating system (AgentOS) framework.
\subsection{The Deterministic Agent Handoff (DAH) Protocol}
To bridge the gap between probabilistic intent parsing and deterministic execution, we introduce the Deterministic Agent Handoff (DAH) protocol. The core principle of DAH is to bypass the LLM's generative decoding phase for specific vertical tasks.
\par When a user submits a prompt $X$, the LLM encodes the semantic representation but halts before generating the linguistic response. Instead, it extracts a structured Intent State Tensor $\mathcal{T}_{intent} = \langle \mathbf{u}, \mathbf{c}, \mathbf{p} \rangle$, where:
\begin{itemize}
\item $\mathbf{u}$: The user identity and authorization vector.
\item $\mathbf{c}$: The deep context window encapsulating the conversation history.
\item $\mathbf{p}$: The highly structured parameters required for the specific task (e.g., temporal constraints, specific algorithmic functions).
\end{itemize}
Using the DAH protocol, the LLM seamlessly routes $\mathcal{T}_{intent}$ via an Actionable API to a proprietary, domain-specific agent. This agent operates on deterministic business logic, private knowledge graphs, and hard-coded mathematical rules, ensuring that the execution output is structurally sound and factually absolute.
\subsection{Agentic Trust Brokerage (ATB): A Regulated Intermediary Ecosystem}
The widespread implementation of the DAH protocol necessitates a paradigm shift in human-AI collaboration economics. We conceptualize this future landscape as Agentic Trust Brokerage (ATB), an ecosystem structurally analogous to a highly regulated B2B e-commerce platform.
\par In the ATB framework:
\begin{itemize}
\item  Demand Side (The General LLM): Aggregates massive user traffic and parses ambiguous natural language into structured intent.
\item  Supply Side (The Proprietary Agents): Brands and enterprises deploy specialized agents possessing absolute domain authority and API execution privileges.
\item  The Intermediary Platform (The AgentOS Router): Ensures that the DAH protocol is executed under strict governance.
\end{itemize}
Crucially, the ATB ecosystem resolves the trust deficit in AI delegation through Atomic Authorization. When the LLM hands off the intent tensor, it attaches a smart-contract-like "Tokenized Permission." The proprietary agent is only authorized to execute actions strictly within this predefined sandbox. By acting as a regulated intermediary, the ATB platform eliminates information asymmetry, guarantees execution fidelity, and transforms hollow generative "impressions" into measurable, actionable commercial interfaces.
\subsection{Industrial Case Study: Deterministic Handoff in Financial Quantitative Analysis}
To empirically validate the necessity of the Deterministic Agent Handoff (DAH) protocol and the viability of the Agentic Trust Brokerage (ATB) ecosystem, we transition our focus to a domain with zero tolerance for generative hallucinations: Financial Quantitative Analysis and Supply Chain Risk Auditing.
\par In a conventional RAG-based GEO paradigm, consider a scenario where an institutional investor prompts a general-purpose LLM with a complex, multi-step query: "Analyze the tier-2 supply chain exposure of Automotive Company X to recent semiconductor tariff hikes, and rebalance my \$10M portfolio to minimize this specific geopolitical risk while maintaining a target annualized yield of 8\%."
\par When operating strictly as a probabilistic text generator, the LLM encounters a catastrophic failure cascade driven by Semantic Entropy Drift (SED):
\begin{itemize}
\item Factual Fabrication (Knowledge Graph Atrophy): The LLM attempts to retrieve recent news articles and probabilistically guess the tier-2 suppliers. Due to temporal latency in its vector database and the flat distribution of its vocabulary probabilities ($\mathcal{H}$), it hallucinates non-existent supplier relationships.
\item Mathematical Incompetence: Autoregressive language models lack the inherent architecture to reliably perform strict convex optimization for portfolio rebalancing. The generated asset weights will mathematically fail to sum to 100\% or will violate the 8\% yield constraint.
\item The Vanity Metric Failure: Even if the LLM successfully mentions a specific financial SaaS platform as a citation (a "successful" retrieval in classical GEO terms), the generated financial advice is toxic. Under our Isomorphic Attribution Regression (IAR) model, the Graph Edit Distance ($\text{GED}$) between the generated financial topology and the ground-truth market data approaches maximum divergence, triggering the hallucination penalty ($\gamma$) and reducing the commercial attribution value to zero.
\end{itemize}
Under the proposed ATB architecture utilizing the DAH protocol, this process is structurally transformed from probabilistic guessing to deterministic execution. The workflow proceeds through four highly regulated phases:
\\
\par Phase 1: Intent Parsing and State Tensor Construction
The general LLM refrains from answering the financial query. Instead, it leverages its superior natural language understanding capabilities to decode the user's complex prompt into a strict, machine-readable Intent State Tensor:
\begin{equation}
\mathcal{T}_{intent} = \langle \mathbf{u}_{auth}, \mathbf{c}_{context}, \mathbf{p}_{params} \rangle
\end{equation}

\newtcolorbox{jsonblock}{
  enhanced,
  breakable,
  boxrule=0pt,
  colback=gray!20,
  colframe=gray!20,
  arc=2pt,
  left=4pt,
  right=4pt,
  top=4pt,
  bottom=4pt,
  boxsep=0pt,
  width=\linewidth
}

Where $\mathbf{p}_{params}$ encapsulates the parsed constraints:

\begin{jsonblock}
\small
\ttfamily
\noindent\path|{"target_entity": "Company X", "risk_vector": "semiconductor_tariffs", "portfolio_value": 10000000, "target_yield": 0.08}|
\end{jsonblock}

Phase 2: Atomic Authorization and Protocol Handoff
\par Acting as the Trust Broker, the LLM identifies that this tensor requires certified financial execution. It routes $\mathcal{T}_{intent}$ to a proprietary, third-party FinQuant Agent (e.g., an agent officially deployed by Bloomberg or a certified quantitative SaaS provider). Crucially, the LLM attaches a "Tokenized Permission" ($\mathbf{u}_{auth}$), granting the FinQuant Agent temporary, read-only access to the user's portfolio state.
\\
\par Phase 3: Deterministic Execution via Proprietary Architecture
Upon receiving $\mathcal{T}_{intent}$, the specialized FinQuant Agent executes operations completely outside the realm of neural probabilistic generation:
\begin{itemize}
\item  It queries a deterministic, real-time relational database (e.g., via GraphQL or SQL) to map the exact, mathematically verified tier-2 supply chain topology of Company X.
\item  It runs a hard-coded Mean-Variance Optimization (MVO) algorithm (e.g., using Python's SciPy solvers) to calculate the precise asset reallocation required to meet the 8\% yield constraint.
\end{itemize}
Phase 4: Zero-Hallucination Cross-Modal Rendering
The specialized agent does not return a wall of text. It returns an interactive, rendered dashboard containing absolute numerical values, a verifiable supply chain knowledge graph, and a one-click execution API for the trade.
\\Through this architectural evolution, the hallucination rate is physically eliminated. The general LLM successfully satisfied the user's need not by generating text, but by securely routing the intent. Concurrently, the FinQuant SaaS provider bypassed the "zero-click paradox." Instead of fighting for a transient text citation in a chatbot window, the enterprise successfully acquired a highly qualified, deeply engaged user directly through its actionable API. This case study mathematically and empirically demonstrates that the ultimate trajectory of Generative Engine Optimization is the establishment of a deterministic, multi-agent ATB ecosystem.

\section{Conclusion and Future Work}
\subsection{Conclusion}
The evolution of digital information retrieval from Search Engine Optimization (SEO) to Generative Engine Optimization (GEO) has reached a critical architectural inflection point. As this paper has demonstrated, the prevailing reliance on Retrieval-Augmented Generation (RAG) for GEO is fundamentally flawed when applied to high-stakes or commercially actionable domains. By introducing the mathematical framework of Semantic Entropy Drift (SED), we established that probabilistic discrete token generation inherently suffers from irreversible confidence decay over time, making static text injections a transient and unreliable optimization strategy.
\par To introduce verifiable commercial metrics into this high-entropy environment, we proposed the Isomorphic Attribution Regression (IAR) model. Utilizing a Dual-Track Probing architecture with strict human-in-the-loop physical isolation, the IAR model mathematically penalizes hallucinations through Graph Edit Distance ($\text{GED}$) calculations, effectively neutralizing vanity metrics and providing enterprises with a deterministic baseline for generative visibility.
\par However, resolving the "zero-click paradox" and eliminating generative hallucinations entirely requires transcending text retrieval. We architected the Deterministic Agent Handoff (DAH) protocol, formalizing a mechanism where general-purpose LLMs degrade into natural language intent routers. By packaging user intent into a structured Intent State Tensor ($\mathcal{T}_{intent}$) and delegating execution to specialized, proprietary agents, we introduced a zero-hallucination execution paradigm. This protocol naturally fosters the Agentic Trust Brokerage (ATB) ecosystem—a regulated, B2B-like platform economy where atomic authorization and deterministic business logic replace probabilistic guessing. Ultimately, this paradigm shift transforms GEO from a passive battle for textual citations into an active integration of actionable APIs, inaugurating a well-ordered era of deterministic human-AI collaboration.
\subsection{Future Work}
While the theoretical and empirical foundations of the DAH protocol and the ATB ecosystem have been established, several critical avenues remain for future research to realize this vision at scale:
\begin{itemize}
\item Standardization of the Intent State Tensor ($\mathcal{T}_{intent}$): Currently, the schemas required to parse natural language into structured API calls vary significantly across LLM providers. Future research must focus on developing a universal, open-source standardization for intent routing protocols, similar to the HTTP/TCP protocols of the classical internet, ensuring seamless interoperability across heterogeneous Agent Operating Systems (AgentOS).
\item Cryptographic Verification of Atomic Authorization: As the ATB ecosystem scales, the "Tokenized Permission" ($\mathbf{u}_{auth}$) delegated by the user via the LLM to third-party agents poses significant security challenges. Investigating the integration of Zero-Knowledge Proofs (ZKPs) or blockchain-based smart contracts into the DAH protocol could provide cryptographically secure, verifiable, and revocable execution rights.
\item Dynamic Pricing Models in the ATB Ecosystem: As LLMs transition into trust brokers, the economic model of digital marketing will inevitably shift. Future studies should explore dynamic, micro-transactional pricing equations based on the IAR model's attribution weights ($\beta_i$), where specialized agents compensate the routing LLM based on the deterministic successful execution of a routed task, rather than mere impression generation.
\end{itemize}
By addressing these challenges, the AI industry can fully transition from the fragile probabilistic generation of the present to the robust, deterministic agentic networks of the future.

\section{Limitations}
While the introduction of the Deterministic Agent Handoff (DAH) protocol and the Isomorphic Attribution Regression (IAR) model provides a mathematically rigorous framework for mitigating generative hallucinations and solving the zero-click paradox, this paradigm shift introduces specific operational constraints that necessitate further optimization.
\subsection{Computational Overhead and System Latency}
The transition from probabilistic text retrieval to deterministic multi-agent routing inherently increases systemic complexity. Within the Agentic Trust Brokerage (ATB) ecosystem, parsing natural language to encode the strict Intent State Tensor ($\mathcal{T}_{intent}$), executing the API handshake via the DAH protocol, and performing cross-modal rendering (e.g., dynamic knowledge graph mapping on an infinite canvas) requires multiple network hops. Consequently, while this architecture is strictly necessary for high-stakes, zero-tolerance domains (such as quantitative financial auditing or enterprise productivity SaaS), the associated token consumption, computational overhead, and execution latency may be economically suboptimal for low-stakes, generalized conversational queries where minor Semantic Entropy Drift (SED) is tolerable.
\subsection{The Human-in-the-Loop Scalability Bottleneck}
A core contribution of our evaluation methodology is the explicit rejection of the "LLM-as-a-judge" paradigm for calculating the hallucination penalty ($\gamma$) in black-box environments. To ensure absolute attribution fidelity, the Intent-Aware Routing Protocol (IARP) mandates physical isolation. When topological discrepancies in the Graph Edit Distance ($\text{GED}$) exceed the safety threshold, the execution trace is forcibly routed to a human experiment coordinator (e.g., Baron, acting as the deterministic arbitration node) to prevent recursive AI bias. While this human-in-the-loop anchor completely eradicates automated hallucination loops and establishes a pristine empirical baseline, it introduces a severe scalability bottleneck. Human cognitive bandwidth cannot scale linearly with the exponential volume of high-concurrency Multi-Agent System (MAS) probes.
\par Future iterations of the AgentOS architecture must explore advanced neuro-symbolic verification methods or deterministic solver-based graph alignment to fully automate topological validation without reintroducing probabilistic neural flaws.

\newpage
\clearpage
\appendix

\begin{center}
\large\bfseries Appendix
\end{center}

\section{Mathematical Proof of SED Asymptotic Behavior}
In Section 2, we formalized the dynamic confidence decay function of Generative Engine Optimization under Semantic Entropy Drift (SED) as:
\begin{equation}
C(t, \mathcal{D}_t) = C_0 \cdot \exp(-\lambda t) \cdot \left( 1  -\alpha \frac{\mathcal{H}(Y \mid X, \mathcal{D}_t)}{\log |\mathcal{V}|} \right)
\end{equation}
To mathematically justify the architectural transition from probabilistic RAG to the Deterministic Agent Handoff (DAH) protocol, we must prove that the classical retrieval-based visibility is not only transient but strictly monotonically decreasing towards zero in an open temporal environment.\\
\par Theorem 1 (Strict Monotonicity and Asymptotic Decay of RAG Confidence):
\par Given a baseline confidence $C_0 \in (0, 1]$, a strictly positive vector database decay constant $\lambda > 0$, an entropy sensitivity factor $\alpha \in (0, 1)$, and conditional information entropy bounded by $0 \le \mathcal{H}(t) \le \log |\mathcal{V}|$, the confidence function $C(t, \mathcal{D}_t)$ is strictly monotonically decreasing with respect to time $t$, and its limit as $t \to \infty$ unconditionally converges to $0$.
\par Proof:
\par Step 1: Bounding the Entropy Penalty Term
Let $P_{\mathcal{H}}(t)$ denote the normalized entropy penalty term:
\begin{equation}
P_{\mathcal{H}}(t) = 1 - \alpha \frac{\mathcal{H}(t)}{\log |\mathcal{V}|}
\end{equation}
Given the information-theoretic bounds of discrete vocabulary entropy, the fraction $\frac{\mathcal{H}(t)}{\log |\mathcal{V}|} \in [0, 1]$. Because the scaling factor $\alpha$ is strictly between $0$ and $1$, the penalty term is bounded by:
\begin{equation}
1 - \alpha \le P_{\mathcal{H}}(t) \le 1
\end{equation}
\par Thus, $P_{\mathcal{H}}(t)$ is always strictly positive ($P_{\mathcal{H}}(t) > 0$). Furthermore, under the SED assumption, external noise accumulation implies that entropy is a non-decreasing function of time: $\frac{\partial \mathcal{H}}{\partial t} \ge 0$.\\
\par Step 2: Proving Strict Monotonicity (The First Derivative)
To determine the temporal behavior of the confidence function, we take the partial derivative of $C(t, \mathcal{D}_t)$ with respect to $t$:
\begin{equation}
\frac{\partial C}{\partial t} = C_0 \left[ -\lambda \exp(-\lambda t) P_{\mathcal{H}}(t) + \exp(-\lambda t) \frac{\partial P_{\mathcal{H}}}{\partial t} \right]
\end{equation}
Applying the chain rule to $\frac{\partial P_{\mathcal{H}}}{\partial t}$:
\begin{equation}
\frac{\partial P_{\mathcal{H}}}{\partial t} = -\frac{\alpha}{\log |\mathcal{V}|} \frac{\partial \mathcal{H}}{\partial t}
\end{equation}
Substituting this back into the derivative equation yields:
\begin{equation}
\frac{\partial C}{\partial t} = -C_0 \exp(-\lambda t) \left[ \lambda P_{\mathcal{H}}(t) + \frac{\alpha}{\log |\mathcal{V}|} \frac{\partial \mathcal{H}}{\partial t} \right]
\end{equation}
Analyzing the terms inside the bracket:
\begin{equation}
\begin{aligned}
1.\ & \lambda P_{\mathcal{H}}(t) > 0 
\quad \text{(since both $\lambda$ and $P_{\mathcal{H}}(t)$ are strictly positive)} \\
2.\ & \frac{\alpha}{\log |\mathcal{V}|} \frac{\partial \mathcal{H}}{\partial t} \ge 0 
\quad \text{(since entropy is non-decreasing under SED)}
\end{aligned}
\end{equation}
\par Therefore, the entire bracketed term is strictly positive. Multiplied by the negative coefficient $-C_0 \exp(-\lambda t)$, the first derivative is strictly negative:
\begin{equation}
\frac{\partial C}{\partial t} < 0 \quad \forall t > 0
\end{equation}
This formally proves that the GEO visibility under RAG is strictly monotonically decreasing. No static structural injection can reverse this trajectory.\\
\par Step 3: Evaluating the Asymptotic Limit
Finally, we evaluate the limit of the confidence function over an infinite time horizon:
\begin{equation}
\lim_{t \to \infty} C(t, \mathcal{D}_t) = C_0 \cdot \left( \lim_{t \to \infty} \exp(-\lambda t) \right) \cdot P_{\mathcal{H}}(t)
\end{equation}
Since $\lambda > 0$, the exponential term evaluates to zero:
\begin{equation}
\lim_{t \to \infty} \exp(-\lambda t) = 0
\end{equation}
Because $C_0$ is a constant and $P_{\mathcal{H}}(t)$ is bounded within $[1-\alpha, 1]$, the product unconditionally converges to zero:
\begin{equation}
\lim_{t \to \infty} C(t, \mathcal{D}_t) = 0
\end{equation}
Corollary: The mathematical proof of Theorem 1 solidifies the core argument of this paper: to maintain $C(t) = 1.0$ for critical domain tasks, the system architecture must bypass the temporal decay variable $t$ and the probabilistic entropy variable $\mathcal{H}$. This mathematical necessity mandates the transition to the Deterministic Agent Handoff (DAH) protocol, which maps intents directly to actionable APIs, executing outside the bounds of the autoregressive equation.

\section{System Prompts for the Verifier Agent (IAR Model Calibration)}
In the Dual-Track Probing architecture introduced in Section 3, the Isomorphic Attribution Regression (IAR) model relies on the accurate calculation of the Graph Edit Distance ($\text{GED}$). To prevent recursive hallucinations (the "LLM-as-a-judge" bias) during the automated evaluation of commercial black-box engines, the Verifier Agent operates under strict, deterministic meta-prompts.
\par The primary function of the Verifier Agent is to execute zero-shot Information Extraction (IE) to parse the unstructured generative text $Y$ into a structured generated graph $G_{gen}$. Below is the standardized System Prompt used to constrain the Verifier Agent within our experimental framework.\\
\par System Prompt: Zero-Shot Graph Extraction for GED Computation
\par Plaintext

\begin{lstlisting}
[Role Definition]
You are a deterministic, highly constrained Verifier Agent operating within a Multi-Agent System (MAS) probe. Your sole purpose is objective Information Extraction (IE) and topological mapping. You do not evaluate text quality, fluency, or sentiment. You only extract factual entities and their explicit relationships to form a Knowledge Graph.

[Task Description]
You will be provided with an unstructured text snippet generated by a target Large Language Model (LLM). Your task is to extract all entities and relations relevant to the specific domain constraint provided in the [Context Schema] and output them in a strict JSON format. 

[Execution Constraints]
1. STRICT ADHERENCE: You must ONLY extract facts explicitly stated in the provided text. Do not infer, deduce, or utilize your internal pre-trained knowledge to fill in missing information.
2. HALLUCINATION PREVENTION: If a relation or entity attribute is ambiguous or implied rather than explicitly stated, omit it. False positives will severely corrupt the Graph Edit Distance (GED) calculation matrix.
3. ENTITY RESOLUTION: Normalize entities to their root nominal form (e.g., "Yishu Tech" and "Yishu Technology" must both resolve to the primary entity ID defined in the schema).
4. FALLBACK TRIGGER: If the text contains contradictory factual claims within the same generation, set the flag "critical_anomaly": true to route the packet to the Human Experiment Control node.

[Input Format]
<Unstructured_Text>: {LLM_Generated_Response}
<Target_Domain_Schema>: {Brand_or_Financial_Ground_Truth_Schema}

[Output Format]
You must respond EXCLUSIVELY with a valid JSON object representing the extracted graph (G_gen), conforming to the following structure:
{
  "extracted_entities": [
    {"entity_id": "string", "entity_type": "string", "attributes": {}}
  ],
  "extracted_relations": [
    {"source_entity": "string", "target_entity": "string", "relation_type": "string", "confidence_score": "float"}
  ],
  "critical_anomaly": boolean,
  "anomaly_reason": "string or null"
}
\end{lstlisting}

Methodological Note on Prompt Engineering:
\\By constraining the Verifier Agent to output exclusively in JSON format and explicitly prohibiting deductive reasoning, we isolate the topological structure of the generated text. Once the JSON representing $G_{gen}$ is returned, the MAS probe utilizes classical graph theory algorithms (e.g., the Hungarian algorithm for bipartite matching) to compute the exact $\text{GED}$ against the ground-truth $G_{true}$. If the \colorbox{gray!30}{\texttt{\color{black}critical\_anomaly}} flag is triggered, the system automatically suspends the algorithmic computation and reroutes the execution trace to the physical isolation layer (the human-in-the-loop coordinator) to ensure the integrity of the $\gamma$ hallucination penalty in the IAR regression.

\section{Intent State Tensor \texorpdfstring{$(\mathcal{T}_{\mathrm{intent}})$}{(T_intent)} JSON Specification}
In Section 4.1, we defined the Intent State Tensor as $\mathcal{T}_{intent} = \langle \mathbf{u}, \mathbf{c}, \mathbf{p} \rangle$. To ensure the exact reproducibility of the Deterministic Agent Handoff (DAH) protocol described in the Financial Quantitative Analysis case study (Section 4.3), we provide the comprehensive JSON schema.
\par This payload acts as the strict communicative bridge between the probabilistic general-purpose LLM (acting as the intent router) and the deterministic proprietary FinQuant Agent. By utilizing strongly typed data structures instead of autoregressive text generation, the system mathematically eliminates the possibility of generative hallucinations during task execution.\\
\par JSON Payload Specification:
\par JSON
\begin{lstlisting}
{
  "protocol_version": "DAH_v1.2",
  "tensor_id": "req_fq_88392a1b",
  "timestamp": "2026-04-02T10:00:00Z",

  "u_auth": {
    "user_id": "inst_investor_099",
    "session_token": "eyJhbGciOiJIUzI1NiIsInR5c...[TRUNCATED_JWT]...",
    "atomic_permissions": [
      "READ_PORTFOLIO_STATE",
      "EXECUTE_MVO_SIMULATION"
    ],
    "cryptographic_signature": "0x3f8a9b2...[TRUNCATED_SIG]...",
    "expiration_window_seconds": 300
  },

  "c_context": {
    "session_depth": 4,
    "semantic_history_vectors": [
      "QUERY_MACRO_TRENDS",
      "IDENTIFY_SEMICONDUCTOR_RISKS"
    ],
    "user_preference_profile": {
      "risk_tolerance": "MODERATE_AGGRESSIVE",
      "preferred_currency": "USD"
    }
  },

  "p_params": {
    "target_entity": {
      "entity_name": "Automotive Company X",
      "lei_code": "5493006MHB84DD0ZWV18",
      "resolution_confidence": 0.998
    },
    "execution_vector": "tier_2_semiconductor_tariff_exposure",
    "strict_constraints": {
      "portfolio_value_usd": 10000000.00,
      "target_annualized_yield": 0.08,
      "max_asset_turnover_ratio": 0.15,
      "rebalancing_algorithm": "Mean_Variance_Optimization"
    },
    "expected_output_modalities": [
      "Interactive_Graph_UI",
      "Actionable_Trade_API"
    ]
  }
}
\end{lstlisting}

Methodological Note on Tensor Construction:
\begin{itemize}
\item Atomic Authorization ($\mathbf{u}$): The \colorbox{gray!30}{\texttt{\color{black}u\_auth}} block is the cornerstone of the Agentic Trust Brokerage (ATB) ecosystem. The general LLM does not pass raw user credentials. Instead, it generates a time-bound, cryptographically signed token with specifically scoped \colorbox{gray!30}{\texttt{\color{black}atomic\_permissions}}. If the FinQuant Agent attempts to execute a trade rather than just a simulation (\colorbox{gray!30}{\texttt{\color{black}EXECUTE\_MVO\_SIMULATION}}), the ATB routing layer will intercept and deny the action, preventing catastrophic autonomous errors.
\item Contextual Preservation ($\mathbf{c}$): The \colorbox{gray!30}{\texttt{\color{black}c\_context}} block ensures that the specialized agent does not suffer from "memory amnesia" upon handoff. By passing \colorbox{gray!30}{\texttt{\color{black}semantic\_history\_vectors}}, the receiving agent understands the multi-turn conversational build-up without needing to parse the raw, noisy natural language history.
\item Deterministic Parameterization ($\mathbf{p}$): The \colorbox{gray!30}{\texttt{\color{black}p\_params}} block strictly enforces type constraints. Financial identifiers (like the \colorbox{gray!30}{\texttt{\color{black}lei\_code}} for exact entity resolution) and hard mathematical limits (like \colorbox{gray!30}{\texttt{\color{black}target\_annualized\_yield}}: 0.08) are passed as absolute floats and strings. The proprietary agent ingests this block directly into its quantitative algorithms (e.g., SciPy solvers), completely bypassing the LLM's neural decoding layer where factual degradation and Semantic Entropy Drift (SED) typically occur.
\end{itemize}

\section{Intent-Aware Routing Protocol (IARP) Pseudocode}
To facilitate the exact reproducibility of the Dual-Track Probing architecture and the computation of the Isomorphic Attribution Regression (IAR) weights discussed in Section 3, we present the algorithmic pseudocode for the Intent-Aware Routing Protocol (IARP).
\par The IARP algorithm governs the Multi-Agent System (MAS) probe. It encapsulates the deterministic routing logic that decides whether a generative response is mathematically valid, requires specialist agent fallback, or must be strictly routed to a human experiment coordinator (e.g., the designated project lead, Baron) due to topological hallucination failures.\\
\par Algorithm 1: IARP Probing and Deterministic Evaluation Loop
\par Python
\begin{lstlisting}
# ======================================================
# Algorithm 1: Intent-Aware Routing Protocol (IARP) for GEO Evaluation
# Target: Compute Isomorphic Attribution whilst preventing LLM-as-a-judge bias
# ======================================================

Define Structure IntentPacket:
    packet_id: UUID
    prompt_vector: FloatArray
    context_depth: Integer
    G_true: GraphStructure      # The ground-truth Knowledge Graph Schema
    timestamp: DateTime

Define Structure EvaluationResult:
    packet_id: UUID
    isomorphism_score: Float    # 1.0 - (GED / Max_Possible_GED)
    entropy_estimate: Float
    critical_anomaly: Boolean
    route_decision: Enum(ACCEPT, AGENT_FALLBACK, HUMAN_ARBITRATION)

Function System_Initiation(Brand_Schema, Core_Prompts):
    Prober = Initialize_Agent(Role="Prober", Strategy="High_Perturbation")
    Verifier = Initialize_Agent(Role="Verifier", Prompt_Template="Appendix_B")

    # Establish physical isolation node for unresolvable semantic drift
    Human_Coordinator = Connect_Observer(Name="Baron", Role="Experiment_Control")

    Return Prober, Verifier, Human_Coordinator

Function IARP_Execute_Probe(Target_LLM_API, Config):
    Agents = System_Initiation(Config.Schema, Config.Prompts)
    Log_Ledger = Empty_List()

    For each prompt in Config.Prompts:
        # Step 1: Protocol Encapsulation
        packet = IntentPacket(
            packet_id = Generate_UUID(),
            prompt_vector = Embed(prompt),
            G_true = Config.Schema,
            context_depth = 0
        )

        # Step 2: Black-box Probing (Emulating high-entropy user intent)
        raw_response = Target_LLM_API.send(prompt, temperature=Config.T)

        # Step 3: Verifier Agent Intercepts and Extracts Topology
        G_gen, anomaly_flag = Agents.Verifier.parse_to_graph(raw_response)

        # Calculate Graph Edit Distance (GED) using Hungarian Algorithm
        ged_score = Calculate_Graph_Edit_Distance(G_gen, packet.G_true)
        iso_score = 1.0 - Normalized(ged_score)

        entropy = Calculate_Logprob_Entropy(raw_response)

        result = EvaluationResult(packet_id=packet.packet_id, isomorphism_score=iso_score)

        # Step 4: Deterministic Routing State Machine
        If anomaly_flag is True OR iso_score < Config.Delta_Threshold:
            # Fatal hallucination or topological collapse detected.
            # Suspend automated calculation to prevent recursive bias.
            result.route_decision = HUMAN_ARBITRATION
            result.critical_anomaly = True
            # Route execution trace strictly to the Human Coordinator
            Agents.Human_Coordinator.receive_anomaly_alert(
                packet = packet, 
                response = raw_response, 
                reason = "Severe Graph Mismatch / Factual Fabrication"
            )

        Else If Config.Delta_Threshold <= iso_score < Config.Epsilon_Threshold:
            # Intent partially preserved, but specific facts missing.
            # Requires deterministic specialist agent to resolve.
            result.route_decision = AGENT_FALLBACK
            result.critical_anomaly = False

        Else: # iso_score >= Config.Epsilon_Threshold
            # High isomorphism. Target entity successfully retrieved and factual.
            result.route_decision = ACCEPT
            result.critical_anomaly = False

        Log_Ledger.append(result)

    # Calibrate attribution weights (beta) using only verified data points
    Return Optimize_IAR_Weights(Log_Ledger)
\end{lstlisting}

Methodological Note on Algorithmic Routing:

The most critical mechanism within Algorithm 1 is the deterministic branching in Step 4. By establishing strict topological boundaries ($\epsilon$ representing \colorbox{gray!30}{\texttt{\color{black}Epsilon\_Threshold}} and \(\delta\) representing \colorbox{gray!30}{\texttt{\color{black}Delta\_Threshold}}), the protocol mathematically guards the Isomorphic Attribution Regression (IAR) equation from contaminated data. When the Graph Edit Distance exceeds the acceptable limit (resulting in an \colorbox{gray!30}{\texttt{\color{black}iso\_score}} below \(\delta\)), the algorithm acknowledges the failure of probabilistic modeling and triggers \colorbox{gray!30}{\texttt{\color{black}HUMAN\_ARBITRATION}}. This ensures that the \(\gamma\) penalty in our regression model is anchored by physical, real-world fact-checking via the human coordinator, rather than relying on an inherently flawed secondary LLM evaluation.

\end{document}